\documentclass[lettersize,journal]{IEEEtran}
\usepackage{amsmath,amsfonts}
\usepackage{algorithmic}
\usepackage{algorithm}
\usepackage{array}
\usepackage{textcomp}
\usepackage{stfloats}
\usepackage{url}
\usepackage{verbatim}
\usepackage{graphicx}
\usepackage{cite}
\usepackage{graphicx}
\usepackage[utf8]{inputenc} 
\usepackage[T1]{fontenc}    

\usepackage{xcolor}         
\usepackage{pifont} 

\usepackage{subfigure}
\usepackage{enumitem} 
\usepackage{todonotes}

\usepackage{color}

\begin{document}

\title{Graph-based Online Monitoring of Train Driver States via Facial and Skeletal Features}


\author{Olivia Nocentini, Marta Lagomarsino, Gokhan Solak, Younggeol Cho, Qiyi Tong,  Marta Lorenzini,\\ and Arash Ajoudani
\thanks{The authors are with the Human-Robot Interfaces and Interaction Laboratory, Istituto Italiano di Tecnologia, 
Genoa, Italy.
Qiyi Tong is also with the Ph.D. Program of National Interest in Robotics and Intelligent Machines (DRIM), Università di Genova, Genoa, Italy.}%

\thanks{Corresponding author's email: {\tt\small olivia.nocentini@iit.it}}%


\thanks{Manuscript received Month DD, YYYY; revised Month DD, YYYY.}}

\markboth{Journal of \LaTeX\ Class Files,~Vol.~XX, No.~Y, Month~2025}%
{Nocentini \MakeLowercase{\textit{et al.}}: Graph-based Online Monitoring of Train Driver States}


\maketitle

\begin{abstract}
Driver fatigue poses a significant challenge to railway safety, with traditional systems like the dead-man switch offering limited and basic alertness checks. This study presents an online behavior-based monitoring system utilizing a customised Directed-Graph Neural Network (DGNN) to classify train driver's states into three categories: \textit{alert}, \textit{not alert}, and \textit{pathological}. To optimize input representations for the model, an ablation study was performed, comparing three feature configurations: skeletal-only, facial-only, and a combination of both. Experimental results show that combining facial and skeletal features yields the highest accuracy (80.88\%) in the three-class model, outperforming models using only facial or skeletal features. Furthermore, this combination achieves over 99\% accuracy in the binary alertness classification. 
Additionally, we introduced a novel dataset that, for the first time, incorporates simulated pathological conditions into train driver monitoring, broadening the scope for assessing risks related to fatigue and health. This work represents a step forward in enhancing railway safety through advanced online monitoring using vision-based technologies.

\end{abstract}


\section{Introduction}

Fatigue is a critical safety concern in railway operations, where long shifts 
and repetitive activities can significantly impair a driver's alertness \cite{filtness2017causes}. Despite regulations aimed at ensuring adequate rest for train drivers, fatigue-related incidents remain alarmingly common. According to research by the AA Charitable Trust, one in eight drivers admits to falling asleep at the wheel, while nearly two-fifths have felt so tired that they feared they might nod off \cite{reportDB}. 
The risk is particularly high with modern automated trains operating on night shifts and monotonous routes \cite{avizzano2019real}. While drivers remain essential for supervision, automation reduces their active engagement for long stretches, increasing the likelihood of inattention and drowsiness \cite{fan2022types}. Fatigued drivers react more slowly to hazards and, in extreme cases, may fall asleep. 
Nighttime driving further heightened this risk due to increased drowsiness during late hours {\cite{fan2022types, harma2002effect}}. 

Given these risks, effective fatigue management in railway operations is essential. A traditional safety mechanism in railway operations is the dead-man switch system, which requires the driver to press a pedal periodically to confirm alertness. If the pedal is not pressed within a specified time frame, an alarm is triggered, and the train may come to a halt. While this system ensures a basic level of vigilance, it also imposes a cognitive and physical burden on the driver, potentially leading to discomfort and operational inefficiency \cite{foot2008questions}, \cite{berdal2024towards}.

To address this problem, more advanced technologies for drowsiness detection have been developed, which can be broadly categorized into three main approaches:
\begin{itemize}
\item Vehicle-based systems use sensors to monitor driving performance indicators such as steering behavior and braking patterns \cite{shahverdy2020driver}. 

\item Physiological-based systems utilize wearable sensors to measure physiological signals such as heart rate variability, muscle activity, and brainwaves \cite{peppoloni2016novel}.

\item Behaviour-based systems employ computer vision techniques to analyze driver behavior through facial expressions, eye movement, head position, yawning frequency, and skeleton posture \cite{avizzano2019real}.

\end{itemize}


The first approach primarily relies on vehicle driving data, including steering angle, speed, and acceleration. A notable example is the Attention Assist system introduced by Mercedes-Benz, which indirectly infers driver fatigue by analyzing driving behavior parameters such as steering frequency and lateral deviation of the vehicle\cite{board2016technological}. Subsequently, the Volkswagen Group developed the Fatigue Detection System (FDS), which evaluates fatigue levels based on driving time and dynamic steering angle analysis\cite{lyu2023fatigue}.
However, 
the effectiveness of such methods is limited in railway operations, where drivers interact with buttons and levers intermittently rather than continuously, as with car steering, making it difficult to assess alertness based on control inputs alone. 

In contrast, physiological signal-based methods provide a more direct and accurate assessment of the driver’s physical condition. Electrocardiogram (ECG), electroencephalogram (EEG), electromyography (EMG), and heart rate variability (HRV) have been extensively utilized to detect driver fatigue \cite{lal2001critical}.
Despite their effectiveness, physiological measurement techniques often suffer from practical limitations. Conventional equipment often presents wearability constraints and is highly susceptible to signal noise, while medical-grade devices remain costly and less feasible for large-scale deployment. 

Given these challenges, behaviour-based fatigue detection has emerged as the dominant research direction, offering a more practical and non-intrusive alternative for online driver monitoring.
The academic community has conducted multi-level innovative research in this domain \cite{joshi2020wild, wang2023fatigue, perkins2022challenges, maheswari2022driver, pandey2023dumodds, salem2024drowsiness}. To address the ethical and safety constraints of real-world road experiments, Wang et al. \cite{wang2023fatigue} developed a virtual driving system based on the CARLA simulation platform\cite{dosovitskiy2017carla}. This system integrates a force-feedback device to simulate real driving interactions and employs a monocular camera to capture physiological indicators such as Eye Aspect Ratio (EAR)\cite{soukupova2016eye, cech2016real}, Mouth Aspect Ratio (MAR)\cite{sikander2018driver}, and Percentage of Eyelid Closure Over the Pupil Over Time (PERCLOS)\cite{Dinges1998PerclosAV}. Using these features, a fatigue state classification model based on Random Forest was established, achieving great classification accuracy in experiments. To overcome the limitations of single-view visual perception, Peng et al. \cite{peng2024multi} developed an immersive driving simulator equipped with a three-degree-of-freedom motion platform. By incorporating a 360° panoramic projection and multi-channel haptic feedback, the system enhances the realism of the driving environment. Additionally, a multi-view camera array comprising four RGB sensors was deployed to enable comprehensive driver behavior data acquisition. The collected data was processed using a Graph Convolutional Network (GCN)\cite{kipf2016semi} to extract spatiotemporal correlation features, significantly improving the accuracy of drowsiness detection. 

Notably, most existing studies have focused on passenger cars and freight vehicles, while research on fatigue detection in railway transportation remains relatively limited. However, the conditions and challenges of alertness detection for train drivers differ from those in road transportation. Unlike car drivers, who maintain relatively stable postures with their hands on the steering wheel, train operators are seated on spring-mounted chairs that introduce head oscillations due to track-induced vibrations. They also perform wider head movements, such as scanning long stretches of track, checking side panels and control screens, and monitoring signals outside the cabin. Moreover, train operators have a different frequency beyond the amplitude of the movements compared to drivers who must constantly check their surroundings and keep the car on a straight line.

To detect train driver alertness, Albadawi et al. \cite{albadawi2023real} developed a real-time drowsiness detection system using computer vision and machine learning to monitor operators' eye patterns via a simple webcam. The system analyzes blink patterns based on neuroscience literature to assess alertness levels. It employs machine learning models—specifically, Multilayer Perceptron, Random Forest, and Support Vector Machines (SVM) —to process consecutive video frames. In this paper, the output of the network is represented by two classes (awake or asleep). In another study, the authors proposed an embedded computer vision system to monitor a train driver’s alertness by analyzing gaze and eyelid blinking using a single camera placed in the cockpit \cite{avizzano2019real}. In \cite{magan2022driver}, the authors developed a Driver Assistance System focused on detecting driver drowsiness to prevent traffic accidents. The system analyzes 60-second sequences of facial images to assess alertness. Two approaches were tested:
Recurrent and Convolutional Neural Network (R-CNN) and Deep Learning with Fuzzy Logic to classify if the driver is awake or not awake.
Finally, in  \cite{essahraui2025real}, the authors developed a real-time, non-intrusive driver drowsiness detection (DDD) system using facial analysis and machine learning techniques. The study evaluates various algorithms, including K-Nearest Neighbors (KNN), SVM, convolutional neural networks (CNNs), and advanced computer vision models like YOLOv5, YOLOv8, and Faster R-CNN and output if the person is awake or not.

However, the aforementioned approaches focus exclusively on facial keypoints, neglecting the skeletal component, which provides valuable and stable information on train driver alertness. 
A more recent work, \cite{chen2023research}, introduced a multi-feature fatigue detection method designed for railway dispatchers, combining facial keypoints with body posture analysis to improve detection accuracy, even under challenging conditions such as facial occlusion or angle variations. This approach considers features such as eye closure duration, blinking patterns, yawning frequency, and posture-related behaviors to assess fatigue levels. The model employs an adaptive Bi-LSTM-SVM algorithm to integrate facial and behavioral data, outputting a three-level classification of fatigue.

A notable limitation of the existing studies is that the classification output is often binary—awake/not awake—or narrowly focused on fatigue levels. This fails to account for situations where the driver may feel physically unwell, which is an equally critical factor in ensuring safety and adequate work performance.  More critically, existing approaches overlook the need to detect pathological conditions where a driver may appear physically present, e.g., with their eyes open, but is not actually alert or conscious due to medical issues such as microsleeps, transient ischemic attacks, or other neurological impairments. Unlike drowsiness, which can be mitigated with alerts or short breaks, these conditions require immediate emergency intervention.
Another major challenge in railway operations is the rapid transition in lighting conditions, especially when trains pass through tunnels or move between open tracks and underground sections. The train driver's movement patterns and abrupt light changes reduce the reliability of conventional alertness detection systems, which often rely on head stability and eye-closure tracking.


This paper presents an innovative monitoring system that assesses train driver states using whole-body motion analysis. 
We implement a custom Directed-Graph Neural Network (DGNN) as it is one of the most effective neural networks for human action recognition, leveraging graph theory \cite{shi2019skeleton}.
Both whole-body keypoints and skeletal links are provided as input to the network to classify the driver state into one of three states: \textit{alert}, \textit{not alert}, or \textit{pathological}.
We referred to \cite{GuidelinesReport}, defining the safety-related guidelines for train drivers, and imitated effects due to illness or pathological condition that inhibit the train driver from performing their work.
Distinguishing between the \textit{not alert} and \textit{pathological} states is crucial, as a simple sound can awaken a non-alert driver, whereas a pathological state requires immediate emergency intervention.
To evaluate the impact of facial and skeletal keypoints, we conducted an ablation study and occlusion sensitivity analysis to assess their contributions to the DGNN's performance.
Our system is designed to function under varying lighting conditions, including \textit{full light}, 
\textit{tunnel-induced light-dark transitions}, and \textit{nighttime} scenarios. 
Additionally, we collected a preliminary artificial dataset that includes the three driver states (\textit{alert}, \textit{not alert}, and simulated \textit{pathological}) across the three lighting conditions (\textit{light}, \textit{sudden light/dark}, and \textit{dark}).
We defined three categories of lighting conditions taking inspiration from \cite{othman2022drivermvt} where the authors collected a driver dataset in full-light and dark conditions. We added also the \textit{sudden light/dark} condition to the previous two since in our case, train drivers can also faced lighting changes during their working period when entering tunnels. 
To the best of our knowledge, this is the first dataset to include recorded simulations of potential pathological indicators (such as tremors, ischemic attacks, and loss of consciousness) affecting train drivers, along with the sudden light variations typical of railway operations, providing a crucial resource for developing and validating AI-driven monitoring systems aimed at enhancing railway safety and preventing accidents. 

\begin{figure}[!t]
\centering
\includegraphics[width=\columnwidth]{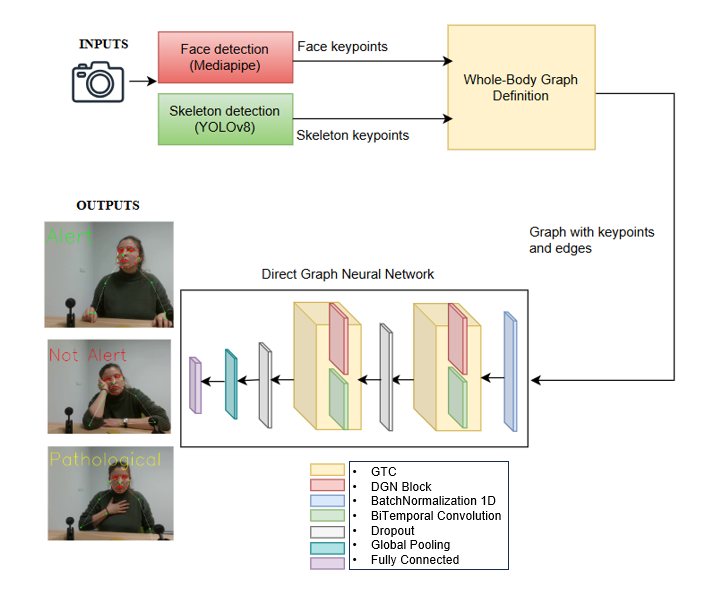}
\caption{Flowchart of our monitoring system framework. The inputs are the images acquired by an RGB camera, from which facial and skeletal features are extracted. These keypoints are used to construct a whole-body graph, which is then processed by a custom DGNN to classify the user's state into three categories: \textit{alert}, \textit{not alert}, or potentially experiencing a \textit{pathological} condition.}
\label{flowchart}
\end{figure}

\section{Methodology}
\label{sec:methodology}
Figure \ref{flowchart} illustrates the overall flowchart of our graph-based monitoring framework for online train driver state assessment, analyzing whole-body motion in RGB images. 
%
Facial and skeletal keypoints are extracted from each video frame to construct a graph. This graph is then processed by a custom Directed-Graph Neural Network (DGNN), which classifies the driver's state into one of three categories: \textit{alert}, \textit{not alert}, or potentially experiencing a \textit{pathological} condition. 

We collected a custom dataset of RGB videos simulating train driver operations under three common railway lighting conditions: \textit{light}, \textit{sudden light/dark}, and \textit{dark} (see Section \ref{sec:sub:dataset} for more details). This dataset enabled us to train the network on more realistic scenarios and conduct an ablation study alongside a model explainability analysis. These analyses aim to understand the relevance of different facial and skeletal keypoints in the network's decision process and assess its sensitivity to keypoint occlusion. 

This paper proposes a rigorous pipeline to address the following research questions (RQs): 
\begin{enumerate}[label=\textbf{RQ\theenumi}.]
    \item \label{RQ1} \textit{
    How effective is a vision and graph-based neural network in detecting the user's alertness and pathological states in simulated railway operations?
    }  
    \item \label{RQ2}  
    \textit{What is the individual and combined contribution of facial and skeleton keypoints to driver state estimation?}  
\end{enumerate}

\subsection{Estimation of facial and skeletal keypoints in video captures and whole-body graph definition}

From each video frame, we estimated the position of the upper body, employing the first 11 keypoints of YOLOv8 \cite{redmon2016you}. 
We took only the upper part of the YOLOv8 detected skeleton since the train driver is usually seated on a chair in front of a dashboard.
Since YOLOv8 provides only two facial keypoints (corresponding to the positions of the left and right eyes), we integrated an additional 35 facial keypoints from Mediapipe's Face Mesh \cite{jakhete2024comprehensive}. This allowed us to better capture facial contours, eyebrows, eye openness, and mouth movements, which YOLOv8 alone cannot detect.
We selected YOLOv8 and Mediapipe Face Mesh for features extraction because both are stable feature detectors for what concerns RGB images \cite{kim2023human}, \cite{dong2024enhanced}.
Moreover, since some keypoints were missing during the extraction process from the RGB videos, instead of setting them to zero, we decided to replace them with the average of the previous 10 samples. This approach aims to make the network more robust.

\begin{figure}[!t]
\centering
\includegraphics[width=1.03\columnwidth]{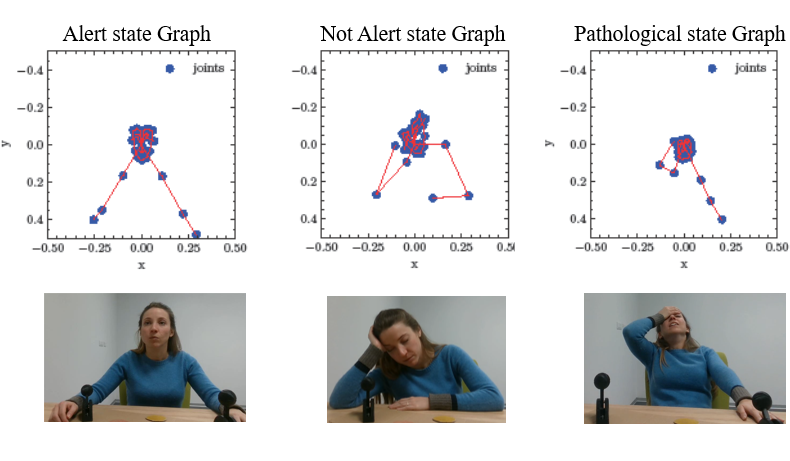}
\caption{Samples of the defined directed graphs, which correspond to the input of our network, in the three tested user's states: \textit{alert}, \textit{not alert}, and \textit{pathological}. The joints are shown in blue, and the bones connecting the joints are represented in red.}
\label{fig:graph}
\end{figure}

Once the keypoints are extracted, we construct a graph to represent the whole-body posture and movements of the user. More specifically, facial and upper-body keypoints serve as nodes, and bones define the edges that connect them. 
The graph is directed, with the starting point defined as the nose. The keypoints were normalized based on the nose position across all three scenarios (face-only, skeleton-only, and face and skeleton keypoints).
In Figure \ref{fig:graph}, samples of the graph defined by connecting the normalized keypoints through edges are reported for the three tested states: \textit{alert}, \textit{not alert}, and the simulated \textit{pathological}. These graphs represent the input of our network.

\subsection{Neural Network for Alert, Not Alert, and Pathological State Classification}
\subsubsection{Network architecture}
The proposed architecture is a customised Directed Graph Neural Network (DGNN) that took inspiration from \cite{shi2019skeleton} designed to process structured spatiotemporal data by jointly modeling graph-based dependencies and temporal dynamics.

The model consists of two Graph-Temporal Convolutional (GTC) blocks, each integrating a Directed Graph Network (DGN) block for relational learning and a Bi-Temporal Convolution (BiTC) layer for temporal feature extraction. Each DGN block updates node and edge features based on directed graph connectivity matrices, 
using learned transformations to aggregate information from incoming and outgoing edges. This operation allows the model to capture structured interactions within the graph. The Bi-Temporal Convolution (BiTC) layer then applies temporal convolutions along the time dimension to both node and edge features, enhancing the model’s ability to learn temporal dependencies.

The input data undergoes an initial batch normalization step, separately for node and edge features, ensuring stable training. The first GTC block expands the feature dimension, while the second block applies strided temporal convolutions (stride=2), reducing the temporal resolution and enabling hierarchical feature extraction. A dropout layer (p=0.7) is applied after each GTC block to mitigate overfitting.

At the final stage, global pooling is applied to the learned node and edge features, followed by a fully connected layer that generates the output predictions. The model employs LeakyReLU activation throughout and uses a Label Smoothing Loss to enhance generalization and prevent overconfidence in predictions.

Our customised model differs from the original implementation of \cite{shi2019skeleton} in several key aspects, aimed at improving performance. First, we reduced the number of convolutional layers, creating a more compact architecture tailored to the relatively small size of our dataset. Second, we introduced dropout layers at specific stages to effectively mitigate overfitting. Third, we replaced certain ReLU activation functions with LeakyReLU, allowing the model to better handle negative values and ensuring greater training stability. Additionally, we incorporated a label smoothing loss function (LabelSmoothingLoss) to enhance robustness when training on noisy datasets, such as ours, which contains inaccuracies in the detected keypoints due to varying lighting conditions during recording.

\subsubsection{Network hyperparameters}
The customised DGNN was trained using the following hyperparameters: a learning rate of 0.001, 60 epochs, a batch size of 1024, Adam as the optimizer, and a lookback (i.e., how many frames the neural network \textit{looks back} into the past) of 15.

\subsection{Relevance of facial and skeletal keypoints in model decision-making}

\subsubsection{Ablation study}
We conducted an ablation study to assess the impact of different categories of keypoints on the performance of the proposed DGNN for train driver's state monitoring. Specifically, we trained three network variants using different input graphs including: 
(1) whole-body keypoints (i.e., combined facial and skeletal keypoints, to examine their synergistic effect); (2) 
facial keypoints only, to determine whether the facial expressions and head movements alone suffice for state estimation, as commonly done in the literature; and (3) skeletal keypoints only, to investigate the importance of upper-body posture and movements in isolation. This approach allowed us to isolate the contributions of facial and skeletal keypoints and determine how their integration enhances the model’s performance.

\subsubsection{Occlusion Sensitivity Analysis}
This analysis aims to identify the most significant keypoints for the functioning of the proposed DGNN and accurate prediction generation. To achieve this, we employed the Occlusion Sensitivity Analysis, a method for understanding the importance of individual features for a deep network's classification. The technique involves occluding (i.e., hiding or removing) one or more keypoints from the input data and observing how this alteration affects the model's performance or predictions. If the occlusion of a specific keypoint causes a significant degradation in the model's predictions, it can be concluded that the keypoint is particularly important for the network's decision-making process. 

\section{EXPERIMENTAL CAMPAIGN}

\label{sec:experiments}

\begin{figure}[!t]
\centering
\includegraphics[width=3in]{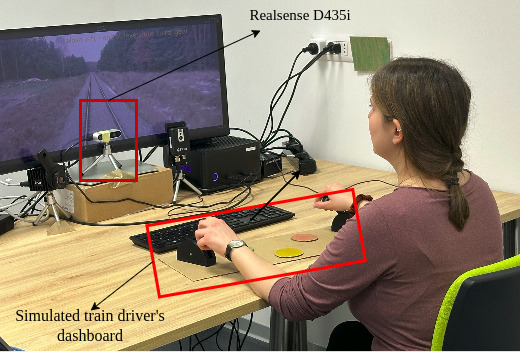}
\caption{The experimental setup consisted of users seated at a desk, following on-screen instructions to manipulate control levers, press dashboard buttons, and check side panels. An RGB camera continuously monitored the subject from the front. We tried to replicate, as closely as possible, the driving cabs configuration of rail vehicles and distances between the train driver and the dashboard specified in \cite{CabinReport}.
}
\label{setup}
\end{figure}

\subsection{Setup}
The experimental setup, depicted in Fig. \ref{setup}, involved participants seated on a chair in front of a table equipped with a computer monitor and a Realsense D435i camera. This camera was positioned directly in front of the participant to capture RGB videos for dataset collection. The sensor operated at a frame rate of 30 fps, producing images with a resolution of 640$\times$480 pixels.
The computer graphic card used to process the data was a Tesla V100-SXM2-16G and the processing steps described in Sec. \ref{sec:methodology} were implemented using Python.

\subsection{
Dataset}
\label{sec:sub:dataset}
The artificial dataset, used for preliminary validation of our framework, consists of videos carried out at Human-Robot Interfaces and Interaction (HRII) Lab, Istituto Italiano di Tecnologia (IIT), in accordance with the Helsinki Declaration. The experimental protocol was approved by the ethics committee Azienda Sanitaria Locale Genovese N.3 (Protocol IIT\_HRII\_ERGOLEAN 156/2020) and participants were recruited from the student body and research staff at IIT. The training set included data from 18 subjects, while the test set comprised data from 2 additional subjects. The dataset features 11 males and 9 females, with ages ranging from 23 to 36 years. Among the participants, 8 subjects wore glasses, while 12 did not.

For each user, we acquired 9 videos of two minutes and half each: the first 3 of them in which the subject is \textit{alert}, the second 3 in which the subject is \textit{not alert}, and the last 3 when the user is pretending to experience a \textit{pathological} condition. Each condition (\textit{alert}, \textit{not alert} and \textit{pathological}) is recorded in 3 different lighting conditions categorized as \textit{light},  \textit{dark}, and \textit{sudden light/dark}, according to the intensity of the light created in the room. 
The \textit{light} condition was defined as the room with 4 lights on, the \textit{dark} condition was defined as the room without light, and the \textit{sudden light/dark} condition was when the lights were turned on/off every 15 seconds to simulate entering/exiting a tunnel.
A video showing the different lighting conditions is provided in the supplementary material of this paper.
During the \textit{alert} phase, the subject was asked to watch a video created by the authors, depicting a train track view. This was designed to help the participant immerse themselves in the role of a train driver. Additionally, instructions appeared on the screen, guiding the participant to perform specific actions, such as moving levers, pressing buttons on the dashboard, or checking side panels. These prompts encouraged head and body movements, enhancing the system’s robustness to variations in head orientation and positioning relative to the camera.
During the \textit{not alert} phase, subjects were free to move as they wished, with the only requirement to keep their eyes closed, pretending to sleep. Finally, during the \textit{pathological} phase, participants simulated experiencing a medical emergency by acting out scenarios such as pretending to faint or clutching their chest, as if suffering from a sudden illness.

\subsection{Analysis}

\subsubsection{Accuracy}
Accuracy during neural network training measures the proportion of correct predictions to the total number of samples, calculated as 
\begin{equation}
\text{Accuracy} = \frac{\text{Number of Correct Predictions}}{\text{Total Number of Samples}} \times 100.
\end{equation}
It evaluates how well the model predicts the true class (ground truth) across the dataset, providing a straightforward metric for performance assessment. Accuracy ranges from 0 to 100\% and is particularly useful for classification tasks with balanced class distributions.

\subsubsection{Confusion Matrices}
We computed the confusion matrices (i.e., a performance evaluation tool comparing predictions to actual labels, highlighting accuracy and misclassification patterns) of the proposed DGNN using whole-body keypoints for state monitoring in three different lighting conditions. The results are reported for both binary classification and three-class classification (\textit{alert}, \textit{not alert} and \textit{pathological}). 

\begin{figure}[!t]
\centering
\includegraphics[width=\linewidth]{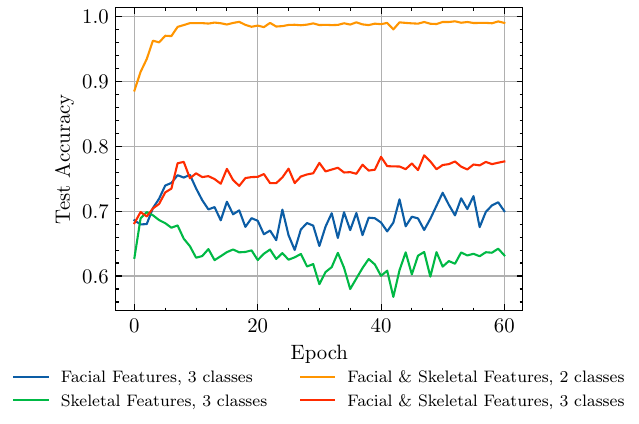}
\caption{Comparison of test accuracies across four network variants, denoted by different colors: 
orange refers to the proposed DGNN for binary state classification using the whole-body graph (considering both skeletal and facial keypoints); 
red corresponds to the proposed three-class DGNN model using the whole-body graph; blue indicates the model using when only facial features, and green represents the model using only skeletal features.}
\label{accuracies}
\end{figure}

\section{Results}
\label{sub:results}
To address the \ref{RQ1} on the effectiveness of a vision and graph-based neural network in detecting user alertness and pathological states in simulated railway operations, we computed the test accuracies of the proposed DGNN model for binary and three-class classification. 
The results using whole-body, facial-only, and skeletal-only keypoints are illustrated in Figure \ref{accuracies}. 
The best performance is observed when only two classes are considered (\textit{alert}/\textit{not alert}), with the accuracy over 99\%. 
The inclusion of the \textit{pathological} class reduces accuracy, but the three-class whole-body model still performs well, outperforming models that rely solely on facial or skeletal keypoints. 


Table \ref{fig:state-of-art} presents a comparison of the performance of our model with previous works, highlighting that our binary model achieves higher overall accuracy under more challenging conditions. 
Specifically, even though Albadawi et al. \cite{albadawi2023real} reached 99\% accuracy with their best model using only facial keypoints, they only tested on a single lighting condition with just two classes. In contrast, our best model acquires 99.36\% accuracy on two classes and covers three different lighting scenarios. We also handled a more complex three-class problem, reaching 80.88\% accuracy, a setting not addressed by prior methods. Furthermore, we combine both face and skeleton information, whereas many previous presents works focus on only one of these modalities.
Moreover, even though \cite{chen2023research} also proposed a three-class model related to fatigue levels, they did not consider the \textit{pathological} condition which is critical according to \cite{GuidelinesReport}. 

In order to address \ref{RQ2}, in Table \ref{table:accuracies_rgb}, test accuracies for the three-class configuration across different lighting conditions are presented. 
As expected, performance is highest in the full \textit{light} condition, followed by the variable \textit{sudden light/dark} condition. In addition, the ablation study  revealed that the face-only model shows a notable drop in the \textit{dark} condition compared to \textit{light} condition, likely due to the reduced visibility and detection reliability of facial keypoints. In contrast, the skeleton-only model maintains similar performance in both \textit{light} and \textit{dark} conditions, indicating that skeletal tracking is probably more robust in low-light conditions. 

\begin{figure}[htbp]
\centering
{\label{fig:a}\includegraphics[width=0.3\textwidth]{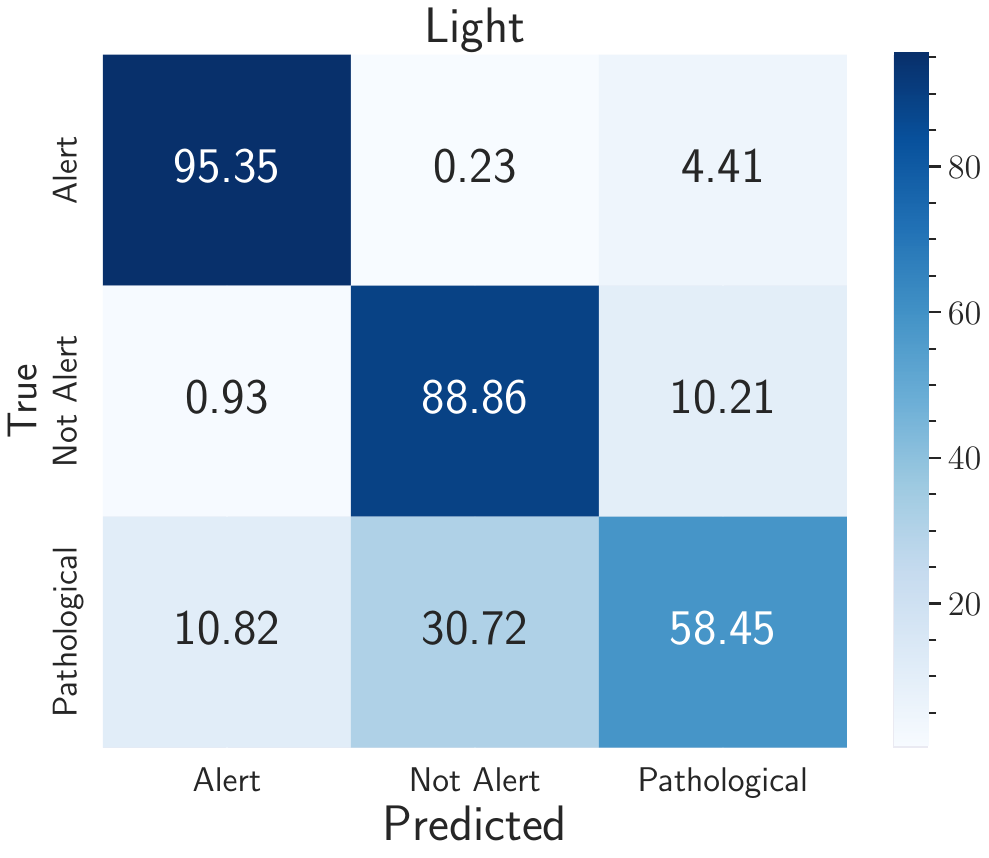}}\qquad
{\label{fig:b}\includegraphics[width=0.3\textwidth]{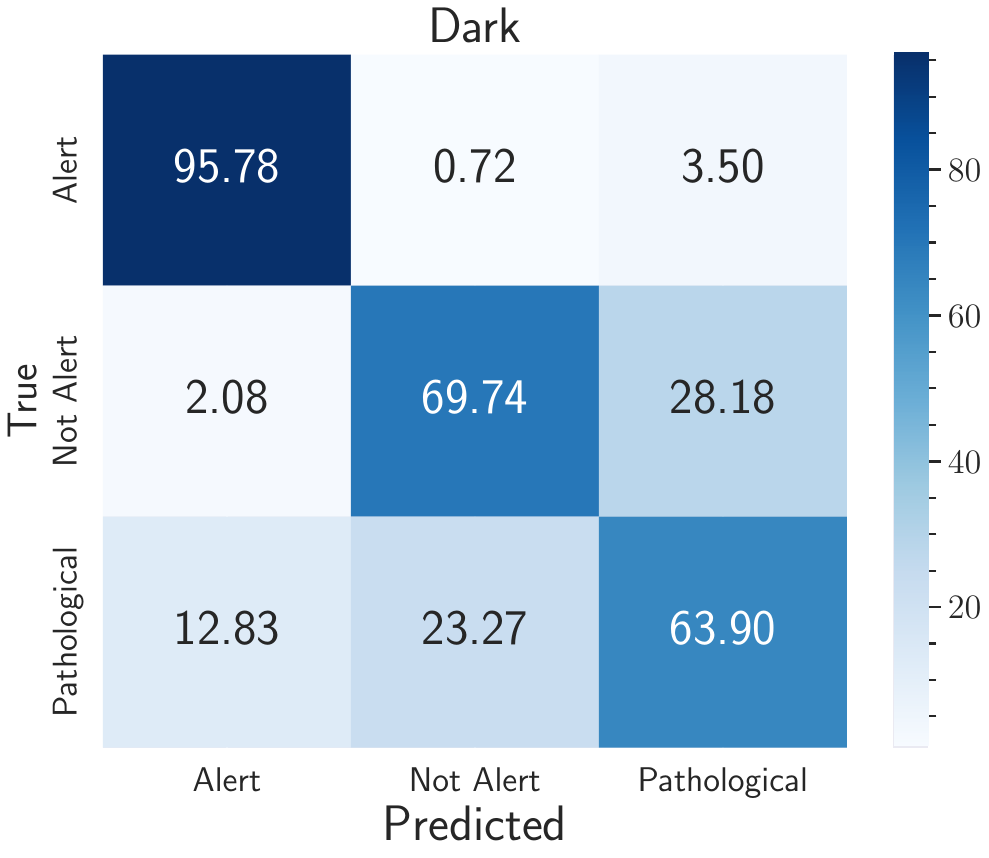}}\\
{\label{fig:c}\includegraphics[width=0.3\textwidth]{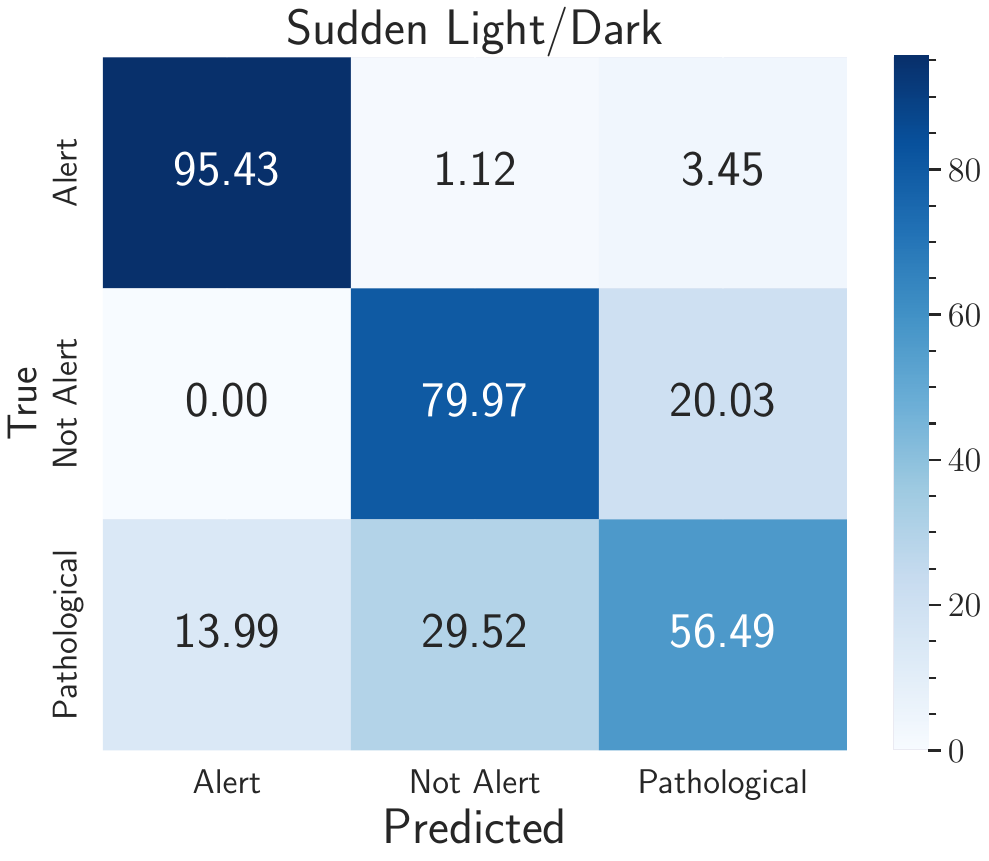}}\qquad%
\caption{Confusion matrices for the proposed three-class model considering whole-body keypoints under the three different lighting conditions.}
\label{fig:cm_rgb}
\end{figure}

To better understand class misclassifications under various lighting conditions, we analyzed the confusion matrices for our three-class model using facial and skeletal keypoints (see Figure \ref{fig:cm_rgb}). The analysis revealed that the \textit{pathological} state is frequently misclassified as the \textit{not alert} condition. 

\begin{figure}
	\centering
	\begin{subfigure}{}
		\includegraphics[width=0.4\textwidth]{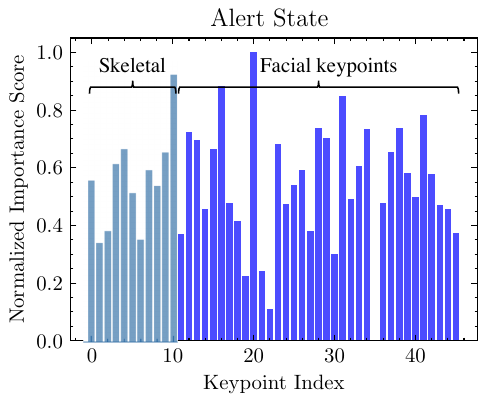}
		\label{fig:subfigA}
	\end{subfigure}
	\begin{subfigure}{}
		\includegraphics[width=0.4\textwidth]{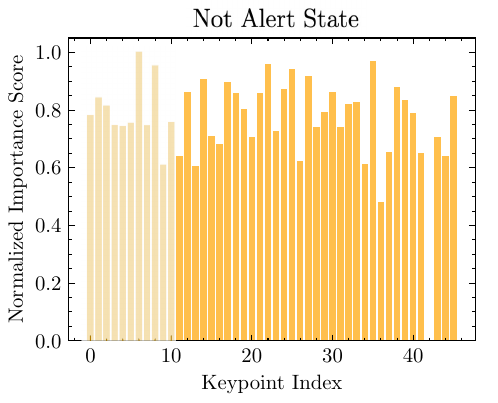}
		\label{fig:subfigB}
	\end{subfigure}
	\begin{subfigure}{}
	        \includegraphics[width=0.4\textwidth]{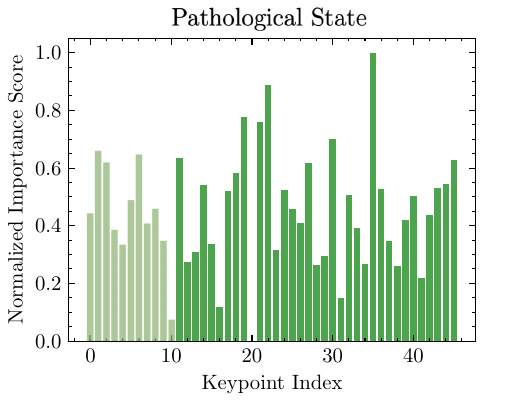}
	        \label{fig:subfigC}
         \end{subfigure}
	\caption{Results of the Occlusion Sensitivity Analysis revealing the significance of individual keypoints to classify the three driver states.}
	\label{fig:oc_rhb}
\end{figure}

Finally, to further address \ref{RQ2} on the individual and combined contribution of facial and skeleton keypoints to driver state estimation, we conducted an Occlusion Sensitivity Analysis on the proposed three-class whole-body model. The results, illustrated in Figure \ref{fig:oc_rhb}, show the relative importance of each keypoint for correct classification. Each bar in the plot represents a keypoint, with its height indicating its significance in distinguishing between the three driver states.

\begin{table*}[]
\caption{Comparison with state-of-the-art methods for train driver state monitoring}
\begin{tabular}{|c|c|c|c|c|c|c|c|}
\hline
\textbf{Authors}      & \textbf{Face} & \textbf{Skeleton} & \textbf{Face \& Skeleton} & \multicolumn{1}{l|}{\textbf{\# Light Conditions}} & \textbf{\# Classes}                                    & \textbf{Algorithm Used}                                                               & \textbf{ Best Accuracy (\%)}                                            \\ \hline
Albadawi et a. 2023 \cite{albadawi2023real}  & x             &                   &                           & 1                                                 & 2                                                      & \begin{tabular}[c]{@{}c@{}}Random Forest\end{tabular} & 99\%                                                              \\ \hline
Avizzano et al. 2019  \cite{avizzano2019real} & x             &                   &                           & 2                                         & 2                                                      & LSTM                                                                                  & -                                                                 \\ \hline
Magan et al. 2022  \cite{magan2022driver}   & x             &                   &                           & 1                                                 & 2                                                      & \begin{tabular}[c]{@{}c@{}}Fuzzy Logic\end{tabular}                           & 93\%                                                              \\ \hline
Essahraui et al. 2025 \cite{essahraui2025real} & x             &                   &                           & 1                                                 & 2                                                      & KNN                                                                                   & 98.89\%                                                           \\ \hline
Chen et al. 2023 \cite{chen2023research}     &               &                   & x                         & 1                                                 & 3                                                      & Bi-LSTM-SVM                                                                           & 96.30\%                                                           \\ \hline
\textbf{Ours}         & \textbf{x}    & \textbf{x}        & \textbf{x}                & \textbf{3}                                        & \textbf{\begin{tabular}[c]{@{}c@{}}2\\ 3\end{tabular}} & \textbf{\begin{tabular}[c]{@{}c@{}}Customised DGNN\\ Customised DGNN\end{tabular}}    & \textbf{\begin{tabular}[c]{@{}c@{}}99.36\%\\ 80.88\%\end{tabular}} \\ \hline
\end{tabular}
\label{fig:state-of-art}
\end{table*}


\begin{table}[]
\centering
\caption{Results of the ablation study: test accuracies of the proposed three-class model across different lighting conditions.}
\begin{tabular}{|c|c|c|c|}
\hline
\textbf{Light condition} & \textbf{Face-only} & \textbf{Skeleton-only} & \textbf{Face and Skeleton} \\ \hline
\textbf{light} & 77.55\%       & 64.44\%           & \textbf{80.88}\%                    \\ \hline
\textbf{dark} & 70.82\%       & 65.59\%           & 76.47\%                    \\ \hline
\textbf{sudden light/dark} & 70.13\%       & 70.55\%           & 77.29\%                    \\ \hline
\end{tabular}
\label{table:accuracies_rgb}
\end{table}


\section{Discussion}
Results of this study highlight that a graph-based neural network using whole-body keypoints analysis was highly effective for binary driver alertness classification and performed well for distinguishing among three classes: \textit{alert}, \textit{not alert} and simulated \textit{pathological} condition. 
Instead, using only skeletal or facial features was insufficient for appropriate three-class classification 
using DGNN, particularly under the varying lighting conditions. 
The ablation study revealed that integrating facial information (e.g., eye and mouth openness, eyebrow arch, and facial contours) with skeletal features results in a richer and more comprehensive evaluation, leading to superior classification performance of the network compared to using either facial or skeletal keypoints alone.

From the confusion matrices presented in Figure \ref{fig:cm_rgb}, we observe that the \textit{pathological} state was sometimes misclassified as the \textit{not alert} state. This misclassification probably occurred because, during data collection, several participants adopted similar whole-body postures in both conditions. For instance, some subjects slumped forward or tilted their heads backwards both when experiencing a pathological event and when falling asleep, making it challenging for the network to distinguish between the two conditions. In the future, we plan to investigate how to better simulate the pathological condition with the help of neuropsychologists.
This tendency explains why the test accuracy of the two-class model was higher compared to the three-class model (see Figure \ref{accuracies}).
Moreover, it can be noticed that the proposed three-class model achieved its highest performance 
in the \textit{light} condition, followed by the \textit{sudden light/dark} condition and the \textit{dark} condition.
This is because RGB cameras work well under well-lit conditions rather than in darkness. Future works will investigate the use of thermal cameras to enhance the performance of the network in dark conditions.

Finally, 
the analysis of individual keypoint importance for the DGNN's decision-making using Occlusion Sensitivity revealed distinct patterns across the three states: \textit{alert}, \textit{not alert}, and \textit{pathological}. In the \textit{alert} state, keypoint importance shows many peaks across both skeletal (YOLOv8) and facial (MediaPipe Face Mesh) keypoints, indicating that both posture and facial expressions contribute to alertness detection. In contrast, the \textit{not alert} state shows a more uniform distribution, with higher reliance on facial keypoints, suggesting that relaxed facial features, such as eyelid drooping and head posture, are key indicators. The \textit{pathological
} state displays a more scattered importance pattern with specific peaks, indicating that the model relies on particular facial features and upper body posture to detect train driver alertness, potentially linked to body weird positions. 

Future works could explore Graph Attention Networks (GATs) or Vision Transformers (ViTs) to enhance features selection by learning an adaptive attention map, focusing on the most discriminative facial and skeletal features for each state. Incorporating these techniques could lead to more robust and explainable human behavior recognition, improving state classification accuracy while reducing reliance on redundant keypoints.

\section{Conclusions}
Driver fatigue remains a critical challenge in railway safety, yet traditional monitoring systems, such as the dead-man switch, offer only rudimentary alertness checks. In this study, we proposed an online monitoring system leveraging a customised Directed-Graph Neural Network (DGNN) and whole-body movement analysis to classify train driver states into \textit{alert}, \textit{not alert}, and \textit{pathological} conditions. 
To determine the most effective input representation for the network, we conducted an ablation study comparing three different keypoint configurations: skeletal-only, facial-only, and whole-body graph. Results indicate that the whole-body graph yields the highest performance, achieving a test accuracy of 80.88\% for the three-class classification in the \textit{light} condition. 
For the two-class classification, the accuracy exceeds 99\%, even under more challenging lighting conditions compared to those considered in previous studies. 

Furthermore, we introduced a novel synthetic dataset that, for the first time, includes the simulation of possible pathological states during train driver activity, thereby expanding the scope of fatigue and health-related risk assessment monitoring.
In this study, we used only RGB videos, as prolonged exposure to infrared cameras may pose a long-term risk to train drivers' eyes. Future work will explore driver state detection using thermal imaging. Moreover, we will leverage GATs or ViTs to enhance feature selection through adaptive attention maps and focus on the most relevant facial and body keypoints. This approach could improve classification accuracy, provide more interpretable human behavior recognition, and reduce reliance on redundant keypoints.




\section*{Acknowledgments}
This work was supported by the Italian Workers’ Compensation Authority INAIL within the VIVA project. 



\bibliography{ref} 
\bibliographystyle{IEEEtran}

\newpage

\end{document}